\documentclass[10pt,twocolumn,letterpaper]{article}

\usepackage{iccv}
\usepackage{times}
\usepackage{epsfig}
\usepackage{graphicx}
\usepackage{amsmath}
\usepackage{amssymb}
\usepackage{booktabs}
\usepackage{multirow}
\usepackage{subfloat}
\usepackage{array}
\usepackage{float}
\usepackage{color}
\usepackage{comment}
\usepackage{enumitem}
\usepackage{caption}
\usepackage{subcaption}
\usepackage{colortbl}
\usepackage{wrapfig}
\usepackage{algorithm}
\usepackage{listings}

\usepackage[pagebackref=false,breaklinks=true,colorlinks,urlcolor=blue,citecolor=blue,linkcolor=blue,bookmarks=false]{hyperref}
\newcolumntype{x}[1]{>{\centering\arraybackslash}p{#1pt}}
\newcolumntype{k}[1]{>{\raggedright\arraybackslash}p{#1pt}}
\newlength\savewidth\newcommand\shline{\noalign{\global\savewidth\arrayrulewidth
		\global\arrayrulewidth 1pt}\hline\noalign{\global\arrayrulewidth\savewidth}}
\usepackage{pifont}

\definecolor{baselinecolor}{gray}{.9}

\usepackage{etoolbox}
\makeatletter
\AfterEndEnvironment{algorithm}{\let\@algcomment\relax}
\AtEndEnvironment{algorithm}{\kern2pt\hrule\relax\vskip3pt\@algcomment}
\let\@algcomment\relax
\newcommand\algcomment[1]{\def\@algcomment{\footnotesize#1}}
\renewcommand\fs@ruled{\def\@fs@cfont{\bfseries}\let\@fs@capt\floatc@ruled
  \def\@fs@pre{\hrule height.8pt depth0pt \kern2pt}%
  \def\@fs@post{}%
  \def\@fs@mid{\kern2pt\hrule\kern2pt}%
  \let\@fs@iftopcapt\iftrue}

\usepackage[capitalize]{cleveref}
\crefname{section}{Sec.}{Secs.}
\Crefname{section}{Section}{Sections}
\Crefname{table}{Table}{Tables}
\crefname{table}{Tab.}{Tabs.}

\iccvfinalcopy 

\def\iccvPaperID{5950} 

\ificcvfinal\fi

\begin{document}

\title{CycleACR: Cycle Modeling of Actor-Context Relations for \\ Video Action Detection}

\author{Lei Chen$^{1}$ \quad
Zhan Tong$^{2}$ \quad
Yibing Song$^{3}$ \quad
Gangshan Wu$^{1}$ \quad
Limin Wang$^{1,4}$\thanks{Corresponding author.}\\
$^{1}$State Key Laboratory for Novel Software Technology, Nanjing University \\
$^{2}$Tencent AI Lab \quad 
$^{3}$AI$^3$ Institute, Fudan University \quad
$^{4}$Shanghai AI Lab \\
\tt\small{leichen1997@outlook.com \quad zhantong.2023@gmail.com
 \quad yibingsong.cv@gmail.com}\vspace{-0.4em} \\ \tt\small{gswu@nju.edu.cn \quad lmwang@nju.edu.cn}
}

\maketitle
\ificcvfinal\thispagestyle{empty}\fi

\begin{abstract}
The relation modeling between actors and scene context advances video action detection where the correlation of multiple actors makes their action recognition challenging. Existing studies model each actor and scene relation to improve action recognition. However, the scene variations and background interference limit the effectiveness of this relation modeling. In this paper, we propose to select actor-related scene context, rather than directly leverage raw video scenario, to improve relation modeling. We develop a Cycle Actor-Context Relation network (CycleACR) where there is a symmetric graph that models the actor and context relations in a bidirectional form. Our CycleACR consists of the Actor-to-Context Reorganization (A2C-R) that collects actor features for context feature reorganizations, and the Context-to-Actor Enhancement (C2A-E) that dynamically utilizes reorganized context features for actor feature enhancement. Compared to existing designs that focus on C2A-E, our CycleACR introduces A2C-R for a more effective relation modeling. This modeling advances our CycleACR to achieve state-of-the-art performance on two popular action detection datasets (i.e., AVA and UCF101-24). We also provide ablation studies and visualizations as well to show how our cycle actor-context relation modeling improves video action detection. Code is available at \href{https://github.com/MCG-NJU/CycleACR}{https://github.com/MCG-NJU/CycleACR}.
\end{abstract}

\section{Introduction}
A fine-grained understanding of sequential video data is to detect actions spatiotemporally. Given an untrimmed video, the spatial-temporal action detection (STAD)~\cite{DBLP:journals/corr/abs-1212-0402,DBLP:conf/cvpr/GuSRVPLVTRSSM18,li2021multisports} aims to localize all the people with their actions recognized. The localization results are represented by bounding boxes in each frame with the associated action labels. Different from general video understanding scenarios (e.g., action recognition~\cite{simonyan2014two,tran2015learning,DBLP:conf/cvpr/CarreiraZ17,DBLP:conf/iccv/Feichtenhofer0M19,tong2022videomae}), the challenge of STAD is that there are multiple people (i.e., actors) in one video with heavy correlations. Such correlations make action detecting of each person difficult. For example, there is a driver and a passenger in a car, which makes detecting the driving action challenging because all these two people appear to sit in the car. Another challenging scenario is that there are a speaker and a listener in a room, which makes detecting the listening action challenging. In order to discriminate different actions between these highly correlated people, modeling the relations between each person and the neighboring scene context has been explored in previous works~\cite{DBLP:conf/eccv/SunSVMSS18,DBLP:conf/cvpr/GirdharCDZ19,DBLP:conf/cvpr/PanCSLS021,DBLP:conf/cvpr/WuF0HKG19,DBLP:conf/cvpr/ZhangTHS19,DBLP:conf/eccv/TangXMPL20}. With the support of scene context, the actions of each person can be better recognized (e.g., detecting the driving action based on the steering wheel) to improve the STAD performance.

\begin{figure}[t]
\centering
\includegraphics[width=1.0\linewidth]{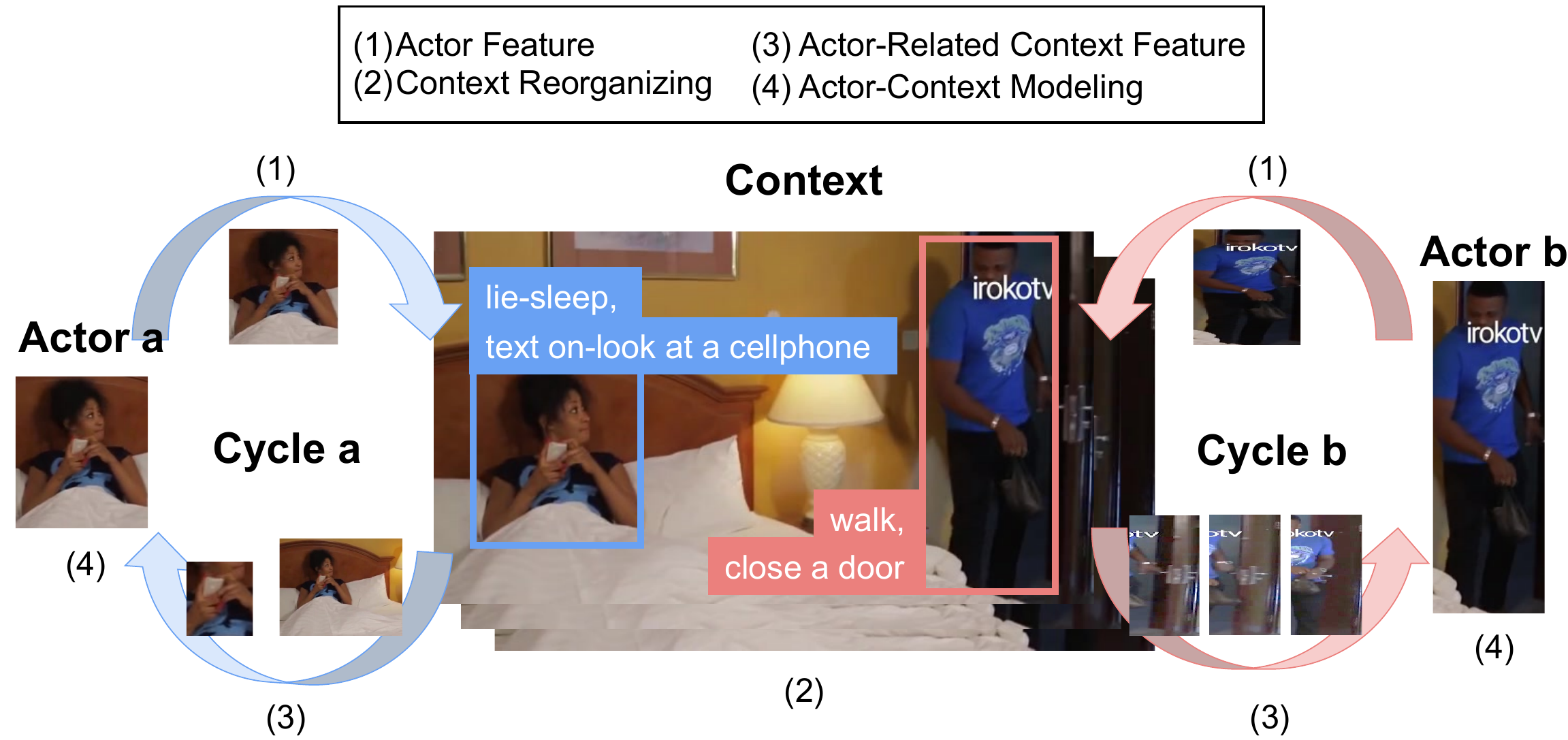}
\vspace{-3mm}
\caption{\textbf{Cycle actor-context relation modeling.} There are two actors in this frame. For each of them, we use a two-step cycle for relation modeling. In the first step (i.e., (1) and (2)), we use actor features to reorganize context features. In the second step (i.e., (3) and (4)), we use actor-related context (e.g., the cellphone and the bed for actor a and the door motion for actor b) for relation modeling.}
\label{fig:intro}
\vspace{-4mm}
\end{figure}

Studies on modeling the relations between actors and context can be mainly categorized into two aspects. First, other than the current actor, the remaining actors are regarded as scene context and involved in actor interactions. Second, the entire scene context features are utilized as the global representation to enhance the actor-centric features. We observe that both two aspects do not well explore the context features to model action relations. For the first one, the scene context other than actors may convey useful features to support action detection (e.g., steering wheel for detecting the driving action). Discarding such context limits relation modeling. For the second one, direct leverage of scene context will bring much irrelevant background feature for relation modeling, which usually overwhelms the relevant one (e.g., car devices other than steering wheel, such as the sun visor, do not help to detect the driving action). Therefore, relation modeling shall be conducted by selecting scene context relevant to one actor. An effective relation modeling will intuitively improve the STAD performance.

In this paper, we propose a Cycle Actor-Context Relation network (CycleACR) to progressively refine actor-relevant scene context and integrate it into video action detection. Fig.~\ref{fig:intro} shows an intuitive view of CycleACR that contains two steps for each actor. In the first step, we connect each actor to the context as shown in (1) and (2). The connection is performed by using actors to reorganize context features by exploring actor-relevant contexts while discarding the irrelevant ones.
Fig.~\ref{fig:intro} shows that the related scene context contains the cellphone and the bed for \textit{actor a}. For \textit{actor b}, the related scene context contains the motions of the door. These actor-relevant contexts are selected and enhanced via cross-attention operations. In the second step, we connect the reorganized context to each actor as shown in (3) and (4). We utilize another cross-attention module for context aggregation. Since the reorganized contexts are actor-related without background interference, the relation modeling is more effective than introducing the global context. The above two steps are summarized as {\em actor-to-context reorganization} and {\em context-to-actor enhancement} in our cycle actor-context relation modeling module. 
Stacking multiple CycleACR modules is able to effectively capture the high-order relation and efficiently exchange useful information between actors an context. 
In addition to utilizing local context features, we correlate each actor feature to long-term temporal scene features from other frames in the memory bank to integrate long-term information as in~\cite{DBLP:conf/cvpr/WuF0HKG19}.

We conduct extensive experiments on the Atomic Visual Actions (AVA) dataset~\cite{DBLP:conf/cvpr/GuSRVPLVTRSSM18} and the UCF101-24 dataset~\cite{DBLP:journals/corr/abs-1212-0402} to verify the effectiveness of our proposed CycleACR. 
On AVA, our CycleACR achieves 34.1\% mAP at IoU 0.5, which outperforms the previous best method ACAR~\cite{DBLP:conf/cvpr/PanCSLS021}. Using all detected boxes produced by an off-the-shelf person detector, it reaches 35.0\% mAP, yielding a new state-of-the-art on AVA validation set. On UCF101-24, it achieves 84.7\% mAP at IoU 0.5, which is better than previous state-of-the-art methods. The superior STAD performance is achieved by effective actor-context relation modeling via our CycleACR, which is further validated in our ablation studies and visualizations. Our CycleACR directs a promising way that mining actor-related scene context improves relation modeling to advance video action detection.

\begin{figure*}[t]
\centering
\includegraphics[width=0.95\linewidth]{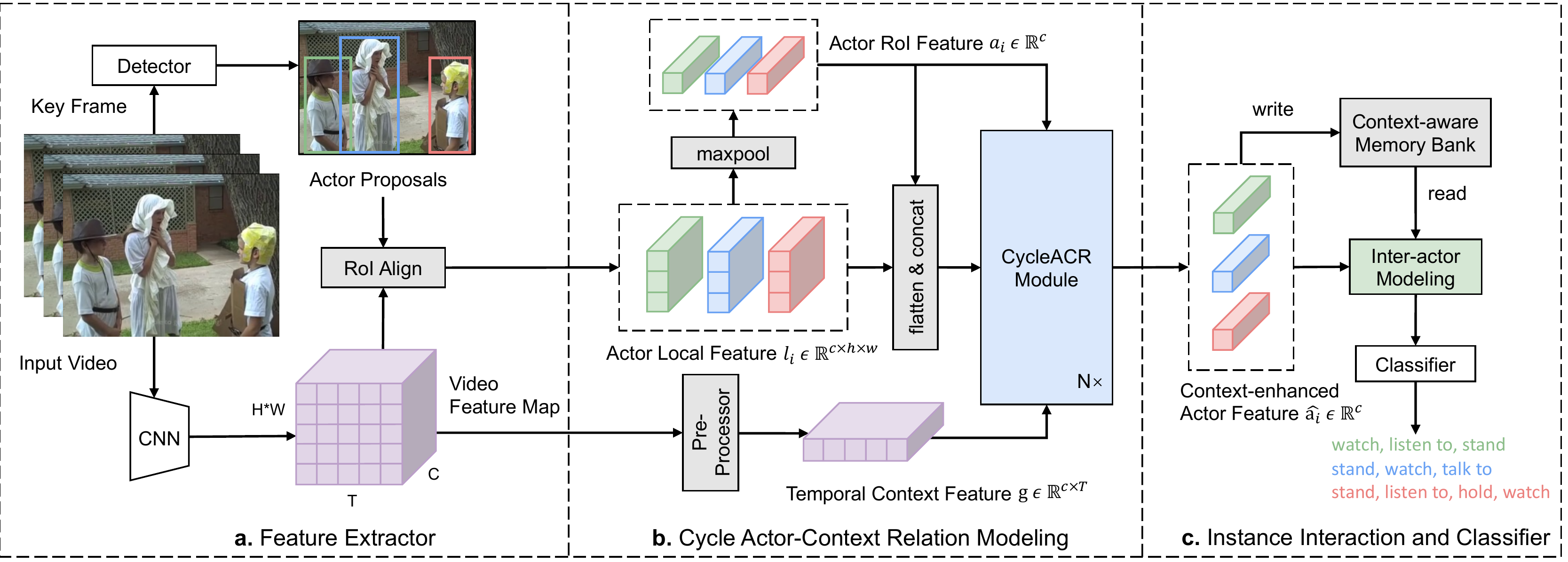}
\vspace{-1mm}
\caption{\textbf{Our CycleACR architecture is composed of three parts.} In (a), we extract video feature maps from a CNN backbone and produce keyframe actor proposals from an off-the-shelf person detector. The actor features are then cropped from the feature map via RoIAlign. In (b), we obtain temporal context features using a pre-processor on the video feature map. Meanwhile, we send actor local features and RoI features to a N-layer CycleACR module for actor-context relation modeling. In (c), the context-enhanced actor features are correlated to other actor features for action detection output.}
\label{fig:pipeline}
\vspace{-4mm}
\end{figure*}

\section{Related Works}
\subsection{Action recognition}
Action recognition is a fundamental task in video understanding and has been widely studied in recent years. Earlier action recognition methods~\cite{wang2016temporal,zhou2018temporal,lin2019tsm,Wang_2021_CVPR} repurpose 2D image convolutional networks (CNNs) and introduce LSTM~\cite{donahue2015long,yue2015beyond} or two-stream networks~\cite{simonyan2014two,feichtenhofer2016convolutional} to learn video temporal structure. Later, various 3D CNNs~\cite{taylor2010convolutional,ji20123d,tran2015learning,varol2017long} are presented to handle video inputs using spatio-temporal kernels.
I3D~\cite{DBLP:conf/cvpr/CarreiraZ17} inflates 2D CNNs into 3D to benefit from ImageNet~\cite{DBLP:conf/cvpr/DengDSLL009} pre-training. Some related works~\cite{qiu2017learning,xie2017rethinking,tran2018closer,diba2018spatio} decouple 3D kernels into separate 2D spatial and 1D temporal kernels to decrease model size. SlowFast~\cite{DBLP:conf/iccv/Feichtenhofer0M19} leverage two pathways to capture fast-changing motion and visual details respectively, and we use it as our video backbone.

\subsection{Spatio-temporal action detection}
Action detection is more challenging than action recognition, requiring not only the prediction of action labels but also the localization of actions in space-time. Most early deep learning methods~\cite{DBLP:conf/aaai/LiLWSQXWX20,DBLP:conf/eccv/PengS16,DBLP:conf/bmvc/SahaSSTC16,DBLP:conf/iccv/SinghSSTC17,DBLP:conf/iccv/WeinzaepfelHS15} follow object detection frameworks~\cite{DBLP:conf/iccv/Girshick15,DBLP:conf/nips/RenHGS15,DBLP:conf/eccv/LiuAESRFB16}, feeding video stream as a series of frames into a 2D CNN network and extracting features separately over each frame for classification. Optical flow is usually introduced to capture motion information. Several recent works~\cite{DBLP:conf/iccv/KalogeitonWFS17a,DBLP:conf/cvpr/YangY0XDK19,DBLP:conf/eccv/LiW0W20,zhao2022tuber} introduce the concept of "tubelet", leveraging the temporal continuity of videos, take as input a sequence of frames and outputs sequences of bounding boxes with associated labels jointly. With the success of 3D CNN in the field of action recognition, many approaches~\cite{DBLP:conf/iccv/HouCS17,DBLP:conf/cvpr/GuSRVPLVTRSSM18,girdhar2018better,DBLP:conf/eccv/SunSVMSS18,DBLP:conf/iccv/Feichtenhofer0M19} use 3D backbones~\cite{DBLP:conf/iccv/TranBFTP15,DBLP:conf/cvpr/CarreiraZ17,DBLP:conf/eccv/XieSHTM18,DBLP:conf/iccv/Feichtenhofer0M19} to extract features, which have significantly improved action detection performance. Moreover, there are emerging methods~\cite{fan2021multiscale,wu2022memvit,tong2022videomae} that leverage transformer-based backbones and achieve promising results.

{\flushleft \bf Relation modeling.}
Modeling of actor and context relation have improved a series of recognition performance including group activity recognition~\cite{gavrilyuk2020actor,yuan2021learning}, human-object interaction detection~\cite{wang2019deep,wang2020contextual} and action recognition~\cite{wang2018videos,herzig2022object}.
With the advent of action detection datasets~\cite{DBLP:conf/cvpr/GuSRVPLVTRSSM18,li2020ava,li2021multisports} in complex multi-person scenes, many recent methods~\cite{DBLP:conf/eccv/SunSVMSS18,DBLP:conf/cvpr/GirdharCDZ19,DBLP:conf/cvpr/WuF0HKG19,DBLP:conf/cvpr/ZhangTHS19,DBLP:conf/wacv/UlutanRS0M20,DBLP:conf/eccv/TangXMPL20,arnab2021unified,DBLP:conf/cvpr/PanCSLS021,faure2023holistic} not only focus on the RoI feature of target actor, but also perform relation modeling with other actor features, object features, and contextual information to deduce the possible action labels of target actor.
Wu {\it et al.}~\cite{DBLP:conf/cvpr/WuF0HKG19} models long-term dependencies between actors via non-local blocks~\cite{wang2018non}.
Girdhar {\it et al.}~\cite{DBLP:conf/cvpr/GirdharCDZ19} employ transformer-style blocks~\cite{DBLP:conf/nips/VaswaniSPUJGKP17} to perform actor and spatio-temporal context interaction. 
Chen {\it et al.}~\cite{chen2021watch} utilize two dynamic instance interactive heads~\cite{DBLP:conf/cvpr/SunZJKXZTLYW021} to implement inter-actor interaction in temporal and spatial dimensions respectively. 
Tang {\it et al.}~\cite{DBLP:conf/eccv/TangXMPL20} design an asynchronously updated bank and integrate multiple types of interactions. 
Pan {\it et al.}~\cite{DBLP:conf/cvpr/PanCSLS021} propose to model the relation between two actors based on their interactions with context.

Our method is a kind of relational modeling, and differs from previous works in that we use a cycle interaction to learn actor-context mutual correlations, rather than a unidirectional actor-context modeling.
It is worth mentioning that compared to AIA~\cite{DBLP:conf/eccv/TangXMPL20}, we do not require an extra object detector and we can interact context information outside detected objects with pre-defined categories. Compared to ACAR~\cite{DBLP:conf/cvpr/PanCSLS021}, we consider the context information as a whole, and do not need to model the relation at each spatial location in the context, which can achieve better performance while saving computation and bank storage.

\section{Proposed Method}
In this section, we will present a detailed description of our proposed CycleACR network on frame-level action detection. The goal is to output the bounding boxes of all actors and their corresponding action labels on the keyframe of the input clip $(\sim 2.5\mathrm{s})$. As shown in Fig.~\ref{fig:pipeline}, the pipeline of our action detection can be divided into three main parts, the second of which is the core module of this paper, CycleACR, as a plug-and-play component that can be integrated into any standard action detection framework.

 \subsection{Feature extractor}
\label{sec:extractor}
Following existing methods~\cite{DBLP:conf/iccv/Feichtenhofer0M19,DBLP:conf/cvpr/WuF0HKG19}, a video backbone network ($e.g.$ SlowFast~\cite{DBLP:conf/iccv/Feichtenhofer0M19}) pre-trained on action recognition extracts the video feature map $f \in \mathbb{R}^{C \times T \times H \times W}$, and $C,T,H,W$ are channel, temporal, height, width respectively. In parallel, an off-the-shelf person detector($e.g.$ Faster R-CNN~\cite{DBLP:conf/nips/RenHGS15}) localizes all actors on the keyframe of the input clip and produces N detected boxes, which are further duplicated along the temporal axis. 3D RoIAlign~\cite{DBLP:conf/iccv/HeGDG17} operation is applied on the video feature map followed by temporal average pooling to calculate N actor local features $l_{i} \in \mathbb{R}^{C \times h \times w}$, $i \in\{1, \cdots, N\}$. 
For follow-up cycle modeling, spatial max pooling is performed on $l_{i}$ to obtain actor RoI feature $a_{i} \in \mathbb{R}^{C}$. For global context feature $g \in \mathbb{R}^{C \times T}$, spatial max pooling is operated on the video feature map as our basic pre-processor to retain video temporal continuity for better relational modeling. To save computational cost, we leverage a $1\times 1$ convolutional layer to reduce the dimension of context feature, actor local and RoI features to $c=1024$ in our implementation.

\subsection{Cycle Actor-Context Relation Modeling}
\label{sec:cycle}
\begin{figure}[t]
\centering
\includegraphics[width=1.0\linewidth]{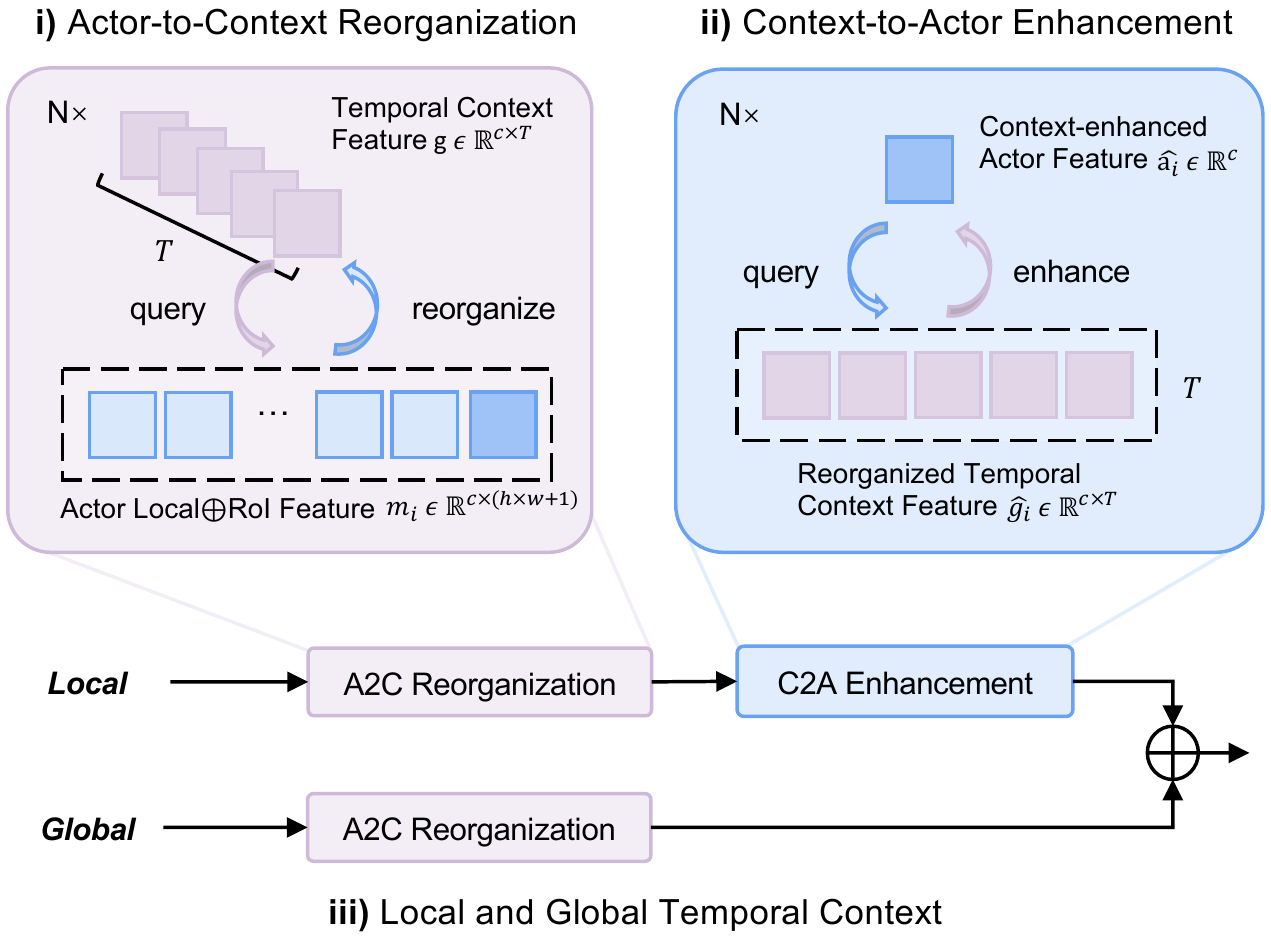}
\vspace{-3mm}
\caption{\textbf{Structure of CycleACR.} We perform actor-context feature interaction in a cycle form. First, we separately reorganize temporal context features according to each actor feature as shown in i). Second, we aggregate the reorganized temporal context to formulate a context-enhanced actor feature as shown in ii). We further conduct the cycle modeling from a local and global temporal perspective as shown in iii).}
\label{fig:detail}
\vspace{-4mm}
\end{figure}

We conduct cycle actor-context relation modeling to efficiently capture the high-order relation and exchange useful information between actors and context.
As shown in Fig.~\ref{fig:detail}, the CycleACR module contains two important elements: (1) {\it Actor-to-Context Reorganization (A2C-R)} use actor local features $l_{i}$ concatenated with actor RoI features $a_{i}$ as the key/value input to select effective context features $g_{i}$ adaptive to each actor based on the shared global context $g$. (2) {\it Context-to-Actor Enhancement (C2A-E)} use the output $\hat{g_{i}}$ from A2C-R as the key/value input and RoI feature $a_{i}$ as the query to aggregate a context-enhanced actor feature $\hat{a_{i}}$. Furthermore, we design local and global branches of cycle modeling to explore temporal-dependent and holistic context information. 

\begin{figure}[t]
\centering
		\begin{center}
		\subfloat[Context-to-actor (C2A) cross-attention module]{
				\includegraphics[width=8cm]{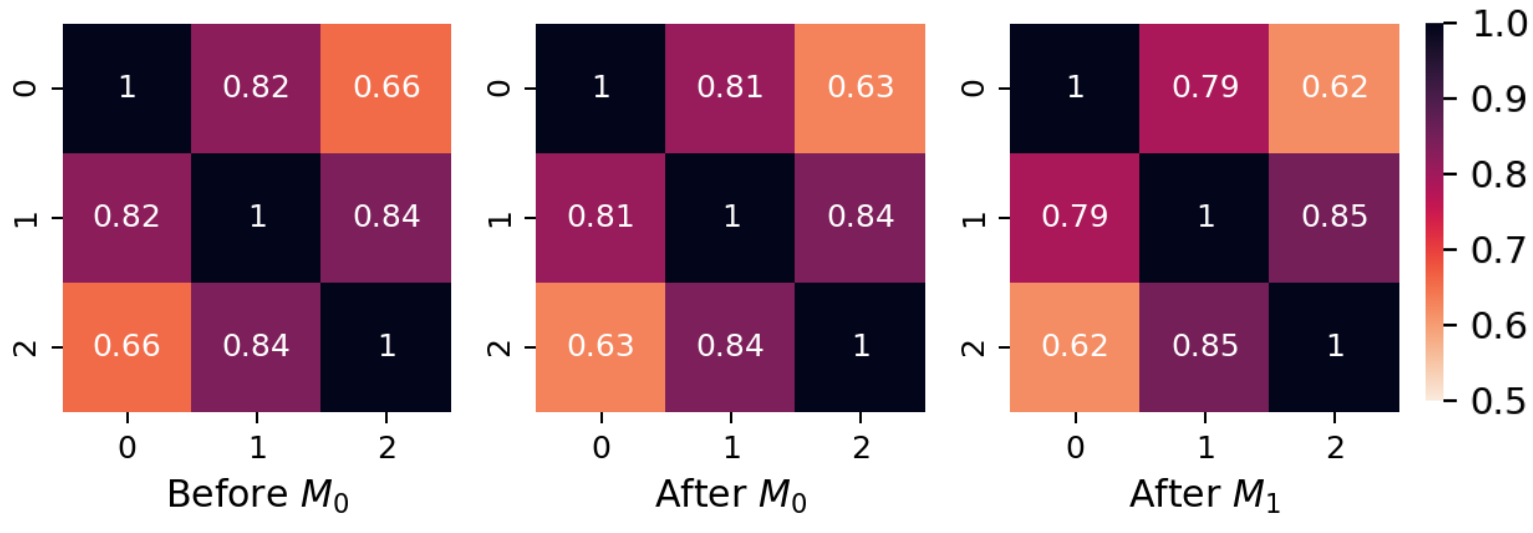}
			\label{fig:cos_sim_1}
		}
        \quad
		\subfloat[Actor-to-context (A2C) cross-attention module]{
				\includegraphics[width=8cm]{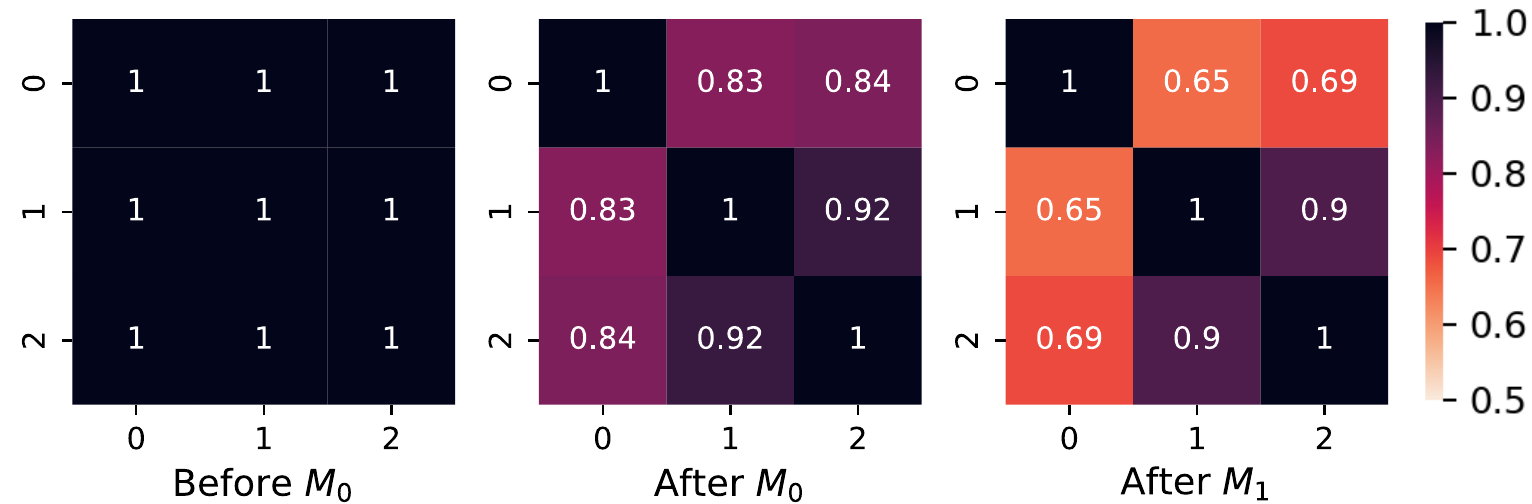}
			\label{fig:cos_sim_2}
		}
		\end{center}
		\vspace{-5mm}
		   \caption{Cosine similarity of three actor features in the same scene. Before $M_{0}$: the input of (a)/(b) attention module is actor RoI feature and global context feature, respectively. After $M_{i}$: the result after applying the $i^{th}$ module.}
        \vspace{-4mm}
\end{figure}

{\flushleft \bf Actor-to-Context Reorganization.}
For action detection in complex scenes, it is often necessary to learn useful information for target actor action classification from contextual information, such as spatio-temporal context cues, explicit object features and other actor features. 
One obvious approach is to use actor features $a_{i} \in \mathbb{R}^{c}$, $i \in\{1, \cdots, N\}$ as the queries and the context feature map $\overline{f} \in \mathbb{R}^{c \times H \times W}$ around the actors as the key/value pairs, termed by context-to-actor cross-attention. In practice, we find that this unidirectional actor-context modeling cannot enlarge the differences between actor features during the process of interaction (see Fig.~\ref{fig:cos_sim_1}). After two layers of C2A cross-attention modules, the similarity among three actor features remains almost constant. It means that the module simply adds a "neutral" piece of global information to each actor, instead of learning their unique information from context (that is, paying different attention to each region of the scene). In other words, we consider it difficult for the model to learn context information adapted to each actor from the amount of interference information.
Instead, we use the global context feature $\overline{g} \in \mathbb{R}^{c}$ as the query to actively learn scene information applicable to each actor. As shown in Fig.~\ref{fig:cos_sim_2}, the contexts corresponding to all actors are the same at the beginning. After two layers of actor-to-context (A2C) attention modules, the context information corresponding to different actors is gradually differentiated, removing the irrelevant noise of the original context. 
Finally, we can obtain reorganized context feature $\hat{\overline{g}_{i}} \in \mathbb{R}^{c}$ for each actor with a degree of discrimination, as a scene prior for follow-up actor-context relation modeling.

In addition, some existing works~\cite{DBLP:conf/eccv/SunSVMSS18,DBLP:conf/cvpr/PanCSLS021} considered the use of spatial context information for relation modeling, while ignoring the temporal information of context, that is, simply average pooling on the temporal dimension. 
We observed that it is also essential to preserve temporal context $g \in \mathbb{R}^{c \times T}$. For example, it can help align the movement trajectories of actors when only having bounding boxes duplicated from keyframes. 
Concretely, actor-to-context cross-attention operates independently on temporal context features to produce frame-level scene priors adaptive to each actor, which are further aggregated by standard context-to-actor cross-attention. In Sec.~\ref{sec:visualization}, we will display the visualization to illustrate that different temporal contexts can represent distinct contributions to action recognition.

We use a variant of non-local blocks as our Actor-to-Context Reorganization module, denoted by A2C-R. In the ablation of Sec~\ref{sec:ablation}, we investigate several other attention mechanisms for comparison and set the non-local one as our default.
In detail, we use temporal context feature $g \in \mathbb{R}^{c \times T}$ as the query of A2C-R. Each actor local feature $l_{i} \in \mathbb{R}^{c \times h \times w}$ flattened to a vector and then concatenated with actor RoI feature $a_{i} \in \mathbb{R}^{c}$ as the memory $m_{i} \in \mathbb{R}^{c \times (h \times w + 1)}$ of A2C-R, which are then projected into key/value pairs. In practice, we find that adding actor RoI feature as a special local feature to the memory can slightly improve the performance. 

By interacting with local information of actor, A2C-R can learn an updated temporal context feature $\hat{g}_{i} \in \mathbb{R}^{c \times T}$ adapt to each actor, which is calculated by:
\begin{align}
& \mathbf{m}_{i}=\operatorname{concat}\left[\mathbf{a}_{i} ; \mathbf{l}_{i,1} ; \mathbf{l}_{i,2} ; \ldots ; \mathbf{l}_{i,h \times w}\right] \\
& \hat{\mathbf{g}_{i}}^{t}=\operatorname{softmax}\left(\frac{\left(\mathbf{W}_{q} \mathbf{g}_{i}^{t}\right)\left(\mathbf{W}_{k} \mathbf{m}_{i}\right)^{\mathrm{T}}}{\sqrt{d}}\right)\left(\mathbf{W}_{v} \mathbf{m}_{i}\right)\\
& \hat{\mathbf{g}_{i}}^{t}=\mathbf{g}_{i}^{t}+\operatorname{dropout}\left(\mathbf{W}_{out}\left(\operatorname{ReLU}\left(\operatorname{norm}\left(\hat{\mathbf{g}_{i}}^{t}\right)\right)\right)\right) \\
& \hat{\mathbf{g}}_{i}=\operatorname{concat}\left[\hat{\mathbf{g}_{i}}^{1} ; \hat{\mathbf{g}_{i}}^{2} ; \ldots ; \hat{\mathbf{g}_{i}}^{T}\right]
\end{align}
where $W_{q}, W_{k}, W_{v} \in \mathbb{R}^{c \times d}$ and $W_{out} \in \mathbb{R}^{d \times c}$ are $1\times 1$ convolutional layers, $d=1024$ in our implementation. We also add layer normalization and dropout for regularization following~\cite{DBLP:conf/cvpr/WuF0HKG19}. So far, we have completed the halfway of our cycle actor-context interaction, from actor to context.

{\flushleft \bf Context-to-Actor Enhancement.}
After getting the actor-related temporal context $\hat{g}_{i} \in \mathbb{R}^{c \times T}$, we regard it as a scene prior and perform standard context-to-actor modeling. This allows actors to learn information from the relevant context that is beneficial for their action recognition, which alleviates the high similarity of learned context features in unidirectional relation modeling.
The second step of CycleACR can also be considered as an aggregation of temporal frame context information. A straightforward way is to perform temporal average pooling on $\hat{g}_{i} \in \mathbb{R}^{c \times T}$ directly, but it is insufficient because some frames are helpful for action recognition while others only bring noise. 

Motivated by the analysis above, we use another modified non-local block with symmetrical inputs as our Context-to-Actor Enhancement module, denoted by C2A-E. It aggregates useful information from the context feature of each frame and generates the final context-enhanced actor feature $\hat{a}_{i} \in \mathbb{R}^{c}$, which can be formulated as:
\begin{align}
& \hat{\mathbf{g}_{i}}^{t}=\hat{\mathbf{g}_{i}}^{t} + \operatorname{pos}\left(t\right), t \in \{1, \cdots, T\}\\
& \hat{\mathbf{a}_{i}}=\operatorname{softmax}\left(\frac{\left(\mathbf{W}_{q} \mathbf{a}_{i}\right)\left(\mathbf{W}_{k} \hat{\mathbf{g}_{i}}\right)^{\mathrm{T}}}{\sqrt{d}}\right)\left(\mathbf{W}_{v} \hat{\mathbf{g}_{i}}\right)\\
& \hat{\mathbf{a}_{i}}=\mathbf{a}_{i}+\operatorname{dropout}\left(\mathbf{W}_{out}\left(\operatorname{ReLU}\left(\operatorname{norm}\left(\hat{\mathbf{a}_{i}}\right)\right)\right)\right)
\end{align}
where $pos$ is a learnable temporal position embedding added to each temporal context for encoding position information, other settings are similar to A2C-R module. In Sec.~\ref{sec:ablation}, we demonstrate that the second step of context-to-actor modeling in CycleACR is an essential one as well.

{\flushleft \bf Local and global temporal context modeling.} 
In the above process, we individually focus on the context information of each frame and aggregate the frame-level scene cues using a C2A-E module. 
Optionally, we can perform a A2C-R module on the global context $\overline{g} \in \mathbb{R}^{c}$ (applied temporal average pooling on $g \in \mathbb{R}^{c \times T}$) to learn a global scene prior for each actor. Compared to local temporal context, global context $\overline{g}$ takes a holistic view of scene’s contribution to action recognition, rather than refining the role of each frame. As in many approaches~\cite{DBLP:conf/iccv/Feichtenhofer0M19,kopuklu2019you,liu2020beyond,zhu2021enriching} that consider modeling from a local and global perspective, we conduct CycleACR modeling of local temporal context $g$ and global temporal context $\overline{g}$ in parallel, as shown in Fig.~\ref{fig:detail} iii). The outputs $\hat{a_{i}}$ and $\hat{\overline{g}_{i}}$ from two branches are concatenated along the channel dimension and then reduced dimension as the final actor feature representation $\hat{a_{i}} \in \mathbb{R}^{c}$.
Note that the updated global context $\hat{\overline{g}_{i}}$ does not require an extra C2A-E for aggregation. In Sec.~\ref{sec:ablation}, we illustrate that the idea of dual local and global context modeling can further improve the performance of action detection.

\subsection{Instance interaction and classifier}
\label{sec:head}
In existing works~\cite{DBLP:conf/cvpr/WuF0HKG19,DBLP:conf/eccv/TangXMPL20,pan2021actor}, target actor features interacts with other actors within the input clip as well as actors in the memory bank with a 60-second window surrounding the clip to capture short-term and long-term inter-actor interactions. Following AIA~\cite{DBLP:conf/eccv/TangXMPL20}, we use an asynchronously updated bank to interact long-term relations without training two separate backbones for bank generation and interactive modeling respectively.

{\flushleft \bf Context-aware memory bank.}
Different from AIA, we put the context-enhanced actor feature $\hat{a_{i}}$ into the bank, which can indirectly store the effective scene information into the bank. When the actor interacts with actors from the bank, it can indirectly see the context of other clips. ACAR~\cite{DBLP:conf/cvpr/PanCSLS021} also uses a bank to store actor-context features, however, since the interaction in ACAR is performed at all spatial locations $(x, y)$, it needs to store each actor-context feature $a_{i} \in \mathbb{R}^{c \times H \times W}$ into the bank for the video. Instead, we only store the context-enhanced actor $\hat{a_{i}} \in \mathbb{R}^{c}$, which achieves the effect of saving scene information without using additional memory overhead. 
For the interactive order, we find that the actor alternately interacts with actors in the current clip and those in the bank resulted in better performance, in consistency with the observations of~\cite{DBLP:conf/eccv/TangXMPL20,wu2022memvit}. 

Finally, the classification layer is applied for multi-label action prediction.

\section{Experiments}
\subsection{Datasets}
AVA~\cite{DBLP:conf/cvpr/GuSRVPLVTRSSM18} is a sparse-annotated spatiotemporal atomic visual action dataset. It consists of 235 training videos and 64 validation videos. Videos are 15-minute segments taken from movies. Bounding boxes are annotated at 1 FPS with possibly multiple labels to generate 56k action instances from 80 atomic actions. Following the standard setting~\cite{DBLP:conf/cvpr/GuSRVPLVTRSSM18}, we evaluate on 60 common classes with mean Average Precision (mAP) as the metric under IoU threshold of 0.5. UCF101-24 is a subset of UCF101~\cite{DBLP:journals/corr/abs-1212-0402} that contains spatiotemporal annotations for 24 action categories in 3,207 videos. Following previous methods~\cite{DBLP:conf/iccv/KalogeitonWFS17a,DBLP:conf/cvpr/YangY0XDK19}, we evaluate frame-mAP with an IoU threshold of 0.5 on the first split.

\subsection{Implementation details}
{\flushleft \bf Person detector.} For AVA, we use the detected person boxes on keyframes from~\cite{DBLP:conf/eccv/TangXMPL20}, which are produced by a Faster R-CNN~\cite{DBLP:conf/nips/RenHGS15} with a ResNeXt-101-FPN backbone in maskrcnn-benchmark~\cite{massa2018mrcnn}. The model is pre-trained on ImageNet~\cite{DBLP:conf/cvpr/DengDSLL009} and MSCOCO~\cite{DBLP:conf/eccv/LinMBHPRDZ14} datasets, and then fine-tuned on AVA. For UCF101-24, we fine-tune the person detector on UCF101-24 with per-frame annotations, which produces a 93.57\% recall on the test set.

{\flushleft \bf Backbone.} State-of-the-art 3D CNN backbone SlowFast~\cite{DBLP:conf/iccv/Feichtenhofer0M19} R50/R101 network pre-trained on Kinetics-700~\cite{DBLP:conf/cvpr/CarreiraZ17} and transformer-based backbone Swin-Tiny/Base~\cite{liu2022video} network pre-trained on Kinetics-400/600 are used to extract video features in our experiments. For SlowFast backbones, we set the spatial stride of res5 to 1 and use a dilation of 2 for its filters following the recipe in~\cite{DBLP:conf/iccv/Feichtenhofer0M19}. The $T \times \tau = 4 \times 16, \alpha = 8$ is used for SlowFast R50, and $T \times \tau = 8 \times 8$, $\alpha = 4$ for SlowFast R101.

{\flushleft \bf Training and inference.} Our model inputs are 32 RGB frames sampled from 64-frame clips with a stride of 2, and the short side of the input frames are scaled to 256 pixels. We train all models in an end-to-end manner using a SGD optimizer with a batch size of 32 clips on 8 GPUs. For SlowFast backbone, the base learning rate is 0.0008 and the head learning rate is 0.008. For Swin backbone, both the base and the head learning rates are 0.0024. We apply a linear warm-up~\cite{DBLP:journals/corr/GoyalDGNWKTJH17} at the first 2k iterations. We also use weight decay of $10^{-7}$ and Nesterov momentum of 0.9. We train models for 50k iterations and reduce the learning rate by 10$\times$ at 30k and 40k iterations. For training, only ground-truth person boxes are used, and random jitter is performed for data augmentation following~\cite{DBLP:conf/eccv/TangXMPL20}. During inference, the detected boxes with confidence $\geq$ 0.8 are used for action classification, and the final action score is the product of box confidence and action category probability. 

For UCF101-24, the base learning rate is 0.0002. We train models for 9k iterations and reduce the learning rate by 10$\times$ at 5k and 7k iterations.
For training, we use ground-truth person boxes as positive samples and detected boxes overlapping with ground-truth boxes by IoU less than 0.3 as negatives samples. The detected boxes with confidence $\geq$ 0.8 are used in training and all the detected boxes are used for inference. Other settings are the same as AVA.

All methods are implemented using AlphAction~\cite{DBLP:conf/eccv/TangXMPL20} and our code will be released to facilitate the future research.

\subsection{Ablation studies}
\label{sec:ablation}
We conduct ablation studies to investigate the effectiveness of each component in our CycleACR network. All results are reported on AVA v2.2 with a SlowFast R101 $8\times 8$ backbone, except for the ablation of backbones.

{\flushleft \bf Interaction mode.}
To demonstrate the importance of learning actor-context mutual correlations, we first compare normal context-to-actor (C2A) cross-attention, the proposed actor-to-context (A2C) and cycle actor-context (CycleACR) cross-attention interaction modes.

\setlength{\tabcolsep}{2pt}
\begin{table}
\small
\begin{center}
\begin{tabular}{k{110}c} 
module & mAP \\
\shline C2A cross-attention & $33.47$ \\
A2C cross-attention & $33.59$ \\
\rowcolor{baselinecolor} CycleACR cross-attention & $34.12$
\end{tabular}
\vspace{-1mm}
\caption{\textbf{Interaction mode.}}
\label{table:mode}
\end{center}
\vspace{-8mm}
\end{table}
\setlength{\tabcolsep}{1.4pt}

As shown in Table~\ref{table:mode}, the A2C mode reorganizes the global context adaptive to each actor, which outperforms the normal C2A mode. The cycle modeling of A2C reorganization and C2A enhancement further improves the mAP.

\setlength{\tabcolsep}{2pt}
\begin{table*}
\small
\centering
    \begin{minipage}[t]{0.22\linewidth}
        \centering
        \begin{subtable}[t]{0.85\linewidth}
            \begin{tabular}{ccc}
            $\mathcal{T}_{local}$ & $\mathcal{T}_{global}$ & mAP \\ 
            \shline & & $29.78$ \\
            \hline$\checkmark$ & & $31.82$ \\
            & $\checkmark$ & $31.82$ \\
            \rowcolor{baselinecolor} $\checkmark$ & $\checkmark$ & $32.57$
            \end{tabular}
            \caption{\textbf{Component analysis}}
            \label{tab:component}
        \end{subtable}
    \end{minipage}
    \begin{minipage}[t]{0.22\linewidth}
        \centering
        \begin{subtable}[t]{0.85\linewidth}
            \begin{tabular}{cccc}
            $\mathcal{T}_{local}$ & $\mathcal{T}_{global}$ & $\mathcal{M}_{c}$ & mAP \\
            \shline & & $\checkmark$ & $32.80$ \\
            \hline $\checkmark$ & & $\checkmark$ & $33.40$ \\
            & $\checkmark$ & $\checkmark$ & $33.59$ \\
            \rowcolor{baselinecolor} $\checkmark$ & $\checkmark$ & $\checkmark$ & $34.12$
            \end{tabular}
            \caption{\textbf{Memory bank}}
            \label{tab:bank}
        \end{subtable}
    \end{minipage}
    \begin{minipage}[t]{0.3\linewidth}
        \centering
        \begin{subtable}[t]{0.85\linewidth}
            \begin{tabular}{lcc}
            module & \#param & mAP \\
            \rowcolor{baselinecolor} \shline Non-local~\cite{wang2018non} & $1.06\times$ & $33.40$ \\ Transformer~\cite{DBLP:conf/nips/VaswaniSPUJGKP17} & $1.13\times$ & $33.25$ \\
            Dynamic conv~\cite{DBLP:conf/cvpr/SunZJKXZTLYW021} & $1.21\times$ & $33.45$\\
            & \\
            \end{tabular}
            \caption{\textbf{Study on A2C-R}}
            \label{tab:a2c}
        \end{subtable}
    \end{minipage}
    \begin{minipage}[t]{0.22\linewidth}
        \centering
        \begin{subtable}[t]{0.75\linewidth}
            \begin{tabular}{lc}
            module & mAP \\
            \shline Avg & $32.88$ \\
            \rowcolor{baselinecolor} Non-local~\cite{wang2018non} & $33.40$ \\
            & \\
            & \\
            \end{tabular}
            \caption{\textbf{Study on C2A-E}}
            \label{tab:c2a}
        \end{subtable}
    \end{minipage}
    \begin{minipage}{0.2\linewidth}
        \centering
        \begin{subtable}[t]{1.\linewidth}
        \begin{tabular}{k{72}c}
            module & mAP \\
            \shline $1 \times \{\mathcal{T}_{local}, \mathcal{T}_{global}\}$ & $33.96$ \\
            \rowcolor{baselinecolor} $2 \times \{\mathcal{T}_{local}, \mathcal{T}_{global}\}$ & $34.12$ \\
            $3 \times \{\mathcal{T}_{local}, \mathcal{T}_{global}\}$ & $33.54$\\
            & \\
        \end{tabular}
        \caption{\textbf{CycleACR depth}}
        \label{tab:depth}
    \end{subtable}
    \end{minipage}
    \hspace{1mm}
    \begin{minipage}{0.3\linewidth}
        \centering
        \begin{subtable}[t]{0.85\linewidth}
        \begin{tabular}{lcc}
            backbone & pretrain & mAP \\
            \shline SFR50~\cite{DBLP:conf/iccv/Feichtenhofer0M19} & K700 & $30.90$ \textcolor[RGB]{0,176,80}{(+$3.6$)} \\
            \rowcolor{baselinecolor} SFR101~\cite{DBLP:conf/iccv/Feichtenhofer0M19} & K700 & $34.12$ \textcolor[RGB]{0,176,80}{(+$4.3$)} \\
            Swin-T~\cite{liu2022video} & K400 & $23.21$ \textcolor[RGB]{0,176,80}{(+$3.7$)} \\
            Swin-B~\cite{liu2022video} & K600 & $28.34$ \textcolor[RGB]{0,176,80}{(+$4.1$)} \\
        \end{tabular}
        \caption{\textbf{Different backbone}}
        \label{tab:diff_model}
    \end{subtable}
    \end{minipage}
    \hspace{2mm}
    \begin{minipage}{0.42\linewidth}
        \centering
        \begin{subtable}[t]{1.\linewidth}
        \begin{tabular}{lccc}
            model & GFLOPs & \#param & mAP\\
            \shline SFR101 (baseline)~\cite{DBLP:conf/iccv/Feichtenhofer0M19} & 120.0 & 65.3 & 29.8\\
            AIA~\cite{DBLP:conf/eccv/TangXMPL20} & 130.8 & 103.8 & 32.3\\
            ACAR-Net~\cite{DBLP:conf/cvpr/PanCSLS021} & 212.0{$^\ddag$} & 107.4{$^\ddag$} & 33.3\\
            \rowcolor{baselinecolor} Ours ($1 \times \{\mathcal{T}_{local}, \mathcal{T}_{global}\}$) & 124.8 & 106.9 & $34.0$
        \end{tabular}
        \caption{\textbf{Comparison with other relation modeling methods}}
        \label{tab:comp}
    \end{subtable}
    \end{minipage}
    \vspace{-2mm}
    \caption{\textbf{Ablation experiments.} We use a SlowFast R101 backbone to perform our ablation study on AVA v2.2. $\mathcal{T}_{local}$ refers to the local branch in cycle modeling, i.e., using local temporal context for mutual interactions, and $\mathcal{T}_{global}$ refers to the use of global context. $\mathcal{M}_{c}$ indicates the introduction of memory banks. The used components are marked with "\checkmark". {$^\ddag$}: the code provided by ACAR-Net does not include the implementation of ACFB, so we use the estimation of~\cite{wu2022memvit} as a lower bound.}
    \vspace{-3mm}
\end{table*} 
\setlength{\tabcolsep}{1.4pt}

{\flushleft \bf Component analysis.}
To verify our design, we ablate the effects of different components in CycleACR network, as shown in Table~\ref{tab:component} and~\ref{tab:bank}. We observe that: 
1) Both local and global branch of cycle modeling enhances the effectiveness of actor-context interactions. 2) $\mathcal{M}_{c}$: the introduction of cycle modeling enables context-enhanced actors stored in the bank, further improving the performance of the model. 3) The complete CycleACR yields the best performance, illustrating the necessity and complementarity of the proposed two branches.

\begin{figure}[t]
\centering
\includegraphics[width=7.5cm]{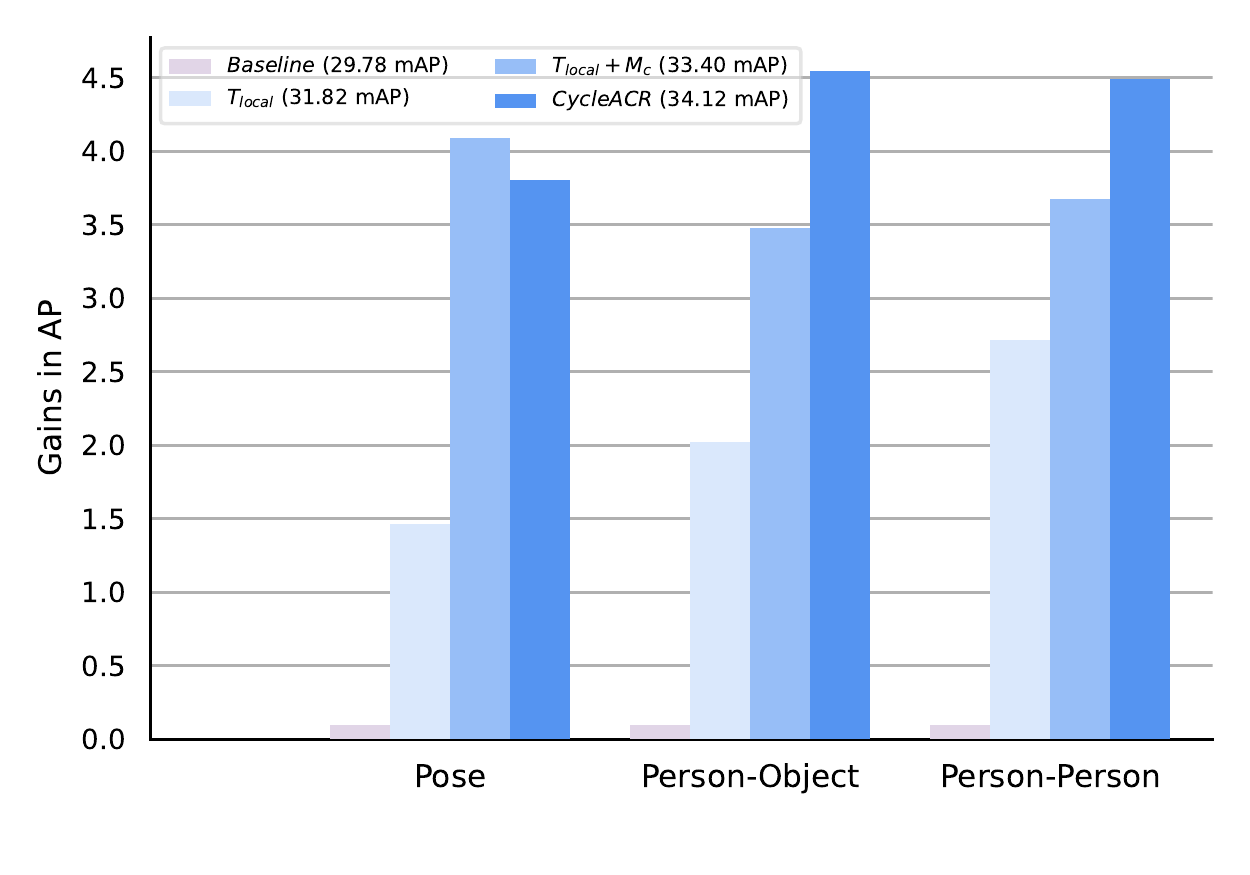}
\vspace{-6mm}
\caption{\textbf{Gains in AP on three major categories of AVA compared to baseline.} $\mathcal{T}_{local}$: introduction of cycle modeling; $\mathcal{T}_{local}+\mathcal{M}_{c}$: cycle modeling combined with context-aware memory bank; $CycleACR$: $\mathcal{T}_{global}+\mathcal{T}_{local}+\mathcal{M}_{c}$.}
\label{fig:3class}
\vspace{-3mm}
\end{figure}

We further break down the performance of different components in CycleACR on three major categories of AVA, as shown in Fig.~\ref{fig:3class}. $\mathcal{T}_{local}$: the implementation of cycle modeling gains significant performance on three categories, especially for "person-object" and "person-person" interactions.
$\mathcal{T}_{local}+\mathcal{M}_{c}$: the introduction of the context-aware memory bank resulted in the largest increase (+2.62) on "pose", which contains most of the motion-dependent categories (e.g., {\it dance}, {\it fall down} and {\it jump}), suggesting that the use of temporal context enables actor motion consistent information to be stored in the bank.
$CycleACR$: the use of local and global branches for context modeling further improves performance on two interactive categories but slightly degrades the performance of pose-related actions. $\mathcal{T}_{global}$ provides a temporal pooling scene information for each actor and sightly reduces the motion information.

{\flushleft \bf A2C-R instantiations.} In Table~\ref{tab:a2c}, we compare several instantiations  of A2C-R. Previous works have used different attention mechanisms to model the interaction of actors with other features, such as modified non-local~\cite{DBLP:conf/cvpr/WuF0HKG19,DBLP:conf/eccv/TangXMPL20}, transformer-style attention~\cite{DBLP:conf/cvpr/GirdharCDZ19,ningperson} and dynamic convolution~\cite{chen2021watch}. Designing a novel block for cycle modeling is not the focus of this paper; here, we ablate the attention mechanisms mentioned above and select the non-local one as our default based on a trade-off between benefits and costs.

{\flushleft \bf C2A-E instantiations.} Table~\ref{tab:c2a} illustrates the necessity of C2A-E, using attention mechanisms to aggregate context information has higher mAP than averaging context directly, i.e., we need normal context-to-actor modeling after obtaining the actor-related context information.

{\flushleft \bf CycleACR depth.} We then experiment on the number of interaction blocks. $N \times \{\mathcal{T}_{local}, \mathcal{T}_{global}\}$ denotes the number of stacking layers in A2C-R/C2A-E used for the local and global branches. The results in Table~\ref{tab:depth} show that stacking two layers in all blocks of CycleACR has the highest mAP, and more than two layers cause performance degradation due to over-fitting, so we use N=2 as our default setting.

{\flushleft \bf Different backbone.} Finally, we ablate different backbones and their model sizes. As presented in Table~\ref{tab:diff_model}, CycleACR can generally improve the performance of each model. 
Surprisingly, CycleACR has a greater improvement on larger models, which indicates that CycleACR works better with more powerful feature representation.

{\flushleft \bf Comparison with other relation modeling methods.} The results in Table~\ref{tab:comp} show that CycleACR achieves better performance with similar or lower computational cost.

\subsection{Comparison with the state-of-the-art}
We compare our CycleACR with existing methods on AVA v2.1 and v2.2 validation set in Table~\ref{table:ava_sota}. Our method surpasses all previous methods in both versions with both video backbones (SFR50 and SFR101). On AVA v2.1, our method achieves 32.9 mAP and outperforms AIA~\cite{DBLP:conf/eccv/TangXMPL20} by 1.7 mAP without an extra object detector. On AVA v2.2, our CycleACR achieves 34.1 mAP without multi-scale and flip testing, surpassing ACAR~\cite{DBLP:conf/cvpr/PanCSLS021} by 0.8 mAP. Note that we do not need to model the relation at each spatial location in the context, saving computational and memory costs. When using all detected human boxes for inference, it reaches 35.0 mAP to establish a new state-of-the-art.

\setlength{\tabcolsep}{2pt}
\begin{table}[t]
\small
\begin{center}
\begin{tabular}{lcccc} 
    \shline model & AVA & $T \times \tau$ & pre-train & mAP \\
    \hline 
    Action Transformer, I3D~\cite{DBLP:conf/cvpr/GirdharCDZ19} & & $64 \times 1$ & K400 & $25.0$ \\
    LFB, R101+NL~\cite{DBLP:conf/cvpr/WuF0HKG19} & & $32 \times 2$ & K400 & $27.7$ \\
    SlowFast, R101~\cite{DBLP:conf/iccv/Feichtenhofer0M19} & & $8 \times 8$ & K600 & $28.2$ \\
    AIA, SFR50~\cite{DBLP:conf/eccv/TangXMPL20} & & $4 \times 16$ & K700 & $28.9$ \\
    AIA, SFR101~\cite{DBLP:conf/eccv/TangXMPL20} & & $8 \times 8$ & K700 & $31.2$ \\
    ACAR-Net, SFR50~\cite{DBLP:conf/cvpr/PanCSLS021} & & $8 \times 8$ & K400 & $28.3$\\
    ACAR-Net, SFR101~\cite{DBLP:conf/cvpr/PanCSLS021} & &  $8 \times 8$ & K400 & $30.0$\\
    TubeR, CSN152~\cite{zhao2022tuber} & & $32 \times 2$ & IG + K400 & $31.7$ \\
    $\mathbf{Ours}$, SFR50 & & $4 \times 16$ & K700 & $\mathbf{29.9}$ \\
    $\mathbf{Ours}$, SFR101 & &  $8 \times 8$ & K700 & $\mathbf{32.9}$ \\
    \hline SlowFast, R101~\cite{DBLP:conf/iccv/Feichtenhofer0M19} & \multirow{9}*{v2.2} & $8 \times 8$ & K600 & $29.1$ \\
    AIA, SFR50~\cite{DBLP:conf/eccv/TangXMPL20} & & $4 \times 16$ & K700 & $29.8$ \\
    AIA, SFR101~\cite{DBLP:conf/eccv/TangXMPL20} & & $8 \times 8$ & K700 & $32.3$ \\
    ACAR-Net, SFR101~\cite{DBLP:conf/cvpr/PanCSLS021} & & $8 \times 8$ & K700 & $33.3$\\
    TubeR, CSN152~\cite{zhao2022tuber} & & $32 \times 2$ & IG + K400 & $33.4$ \\
    $\mathbf{Ours}$, SFR50 & & $4 \times 16$ & K700 & $\mathbf{30.9}$ \\
    $\mathbf{Ours}$, SFR101 & & $8 \times 8$ & K700 & $\mathbf{34.1}$ \\
    $\mathbf{Ours}^{*}$, SFR101 & & $8 \times 8$ & K700 & $\mathbf{35.0}$ \\
    \shline
\end{tabular}
\vspace{-2mm}
\caption{\textbf{Comparison with the state-of-the-art on AVA.} In this table, we display our best results with backbone SlowFast ResNet-50 (SFR50) and ResNet-101 (SFR101). $T \times \tau$ refers to frame number and corresponding sample rate. '*' indicates that the result is tested on all detected boxes. Note that we test all models without multiple scales and flips.}
\label{table:ava_sota}
\end{center}
\vspace{-8mm}
\end{table}
\setlength{\tabcolsep}{1.4pt}

We also compare with previous methods on the UCF101-24 dataset. As shown in Table~\ref{table:ucf_sota}, our CycleACR achieves 84.7 mAP on UCF101-24 test set and outperforms the previous best method, which demonstrates the generalization power of cycle actor-context modeling to new datasets.

\setlength{\tabcolsep}{2pt}
\begin{table}
\small
\begin{center}
\begin{tabular}{lcc} 
\shline model & inputs & mAP \\
\hline T-CNN~\cite{DBLP:conf/iccv/HouCS17} & $\mathrm{V}$ & $67.3$ \\
ACT~\cite{DBLP:conf/iccv/KalogeitonWFS17a} & $\mathrm{V}$ & $69.5$ \\
STEP~\cite{DBLP:conf/cvpr/YangY0XDK19} & $\mathrm{V}+\mathrm{F}$ & $75.0$ \\
I3D~\cite{DBLP:conf/cvpr/GuSRVPLVTRSSM18} & $\mathrm{V}+\mathrm{F}$ & $76.3$ \\
Zhang et al.~\cite{DBLP:conf/cvpr/ZhangTHS19} & $\mathrm{V}$ & $77.9$ \\
MOC~\cite{DBLP:conf/eccv/LiW0W20} & $\mathrm{V}+\mathrm{F}$ & $78.0$ \\
S3D-G~\cite{DBLP:conf/eccv/XieSHTM18} & $\mathrm{V}+\mathrm{F}$ & $78.8$ \\
AIA, R50~\cite{DBLP:conf/eccv/TangXMPL20} & $\mathrm{V}$ & $78.8$ \\
TubeR, CSN152~\cite{zhao2022tuber} & $\mathrm{V}$ & $83.2$ \\ 
ACAR-Net, SFR50~\cite{DBLP:conf/cvpr/PanCSLS021} & $\mathrm{V}$ & $84.3$ \\
\hline $\mathbf{Ours}$ w/o $\mathcal{M}_{c}$, SFR50 & $\mathrm{V}$ & $\mathbf{84.7}$ \\
\shline
\end{tabular}
\vspace{-2mm}
\caption{\textbf{Comparison with the state-of-the-art on UCF101-24.} V is RGB frames and F is optical flows.}
\label{table:ucf_sota}
\end{center}
\vspace{-8mm}
\end{table}
\setlength{\tabcolsep}{1.4pt}

\begin{figure*}[t]
\centering
		\begin{center}
		\subfloat[\textbf{A2C-R attention maps.} Left: the input RGB frame. Right: attention weights of the last layer between the global context and different actors of interest.]{
			\begin{minipage}{5cm}
			\includegraphics[width=1.0\linewidth]{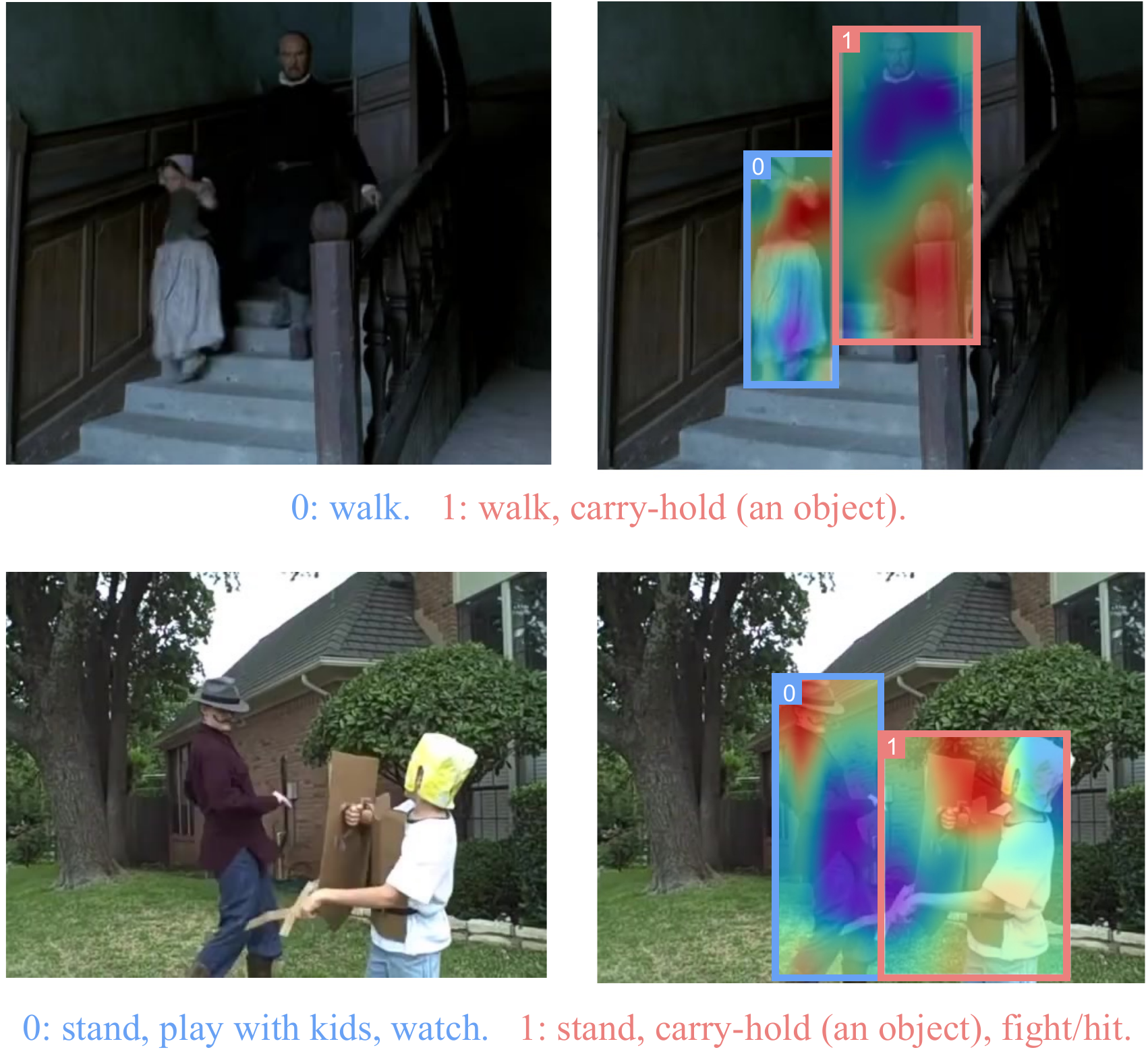}
			\end{minipage}
			\label{fig:roi_attn}
		}
        \quad
		\subfloat[\textbf{C2A-E attention maps.} 1st row: RGB frames of the input clip with a stride of 8 for viewing better, where the third image represents the keyframe. 2nd and 3rd row: attention weights of the $i^{th}$ layer between actors of interest on the keyframe and 32-frame context.]{
			\begin{minipage}{11cm}
		      \includegraphics[width=1.\linewidth]{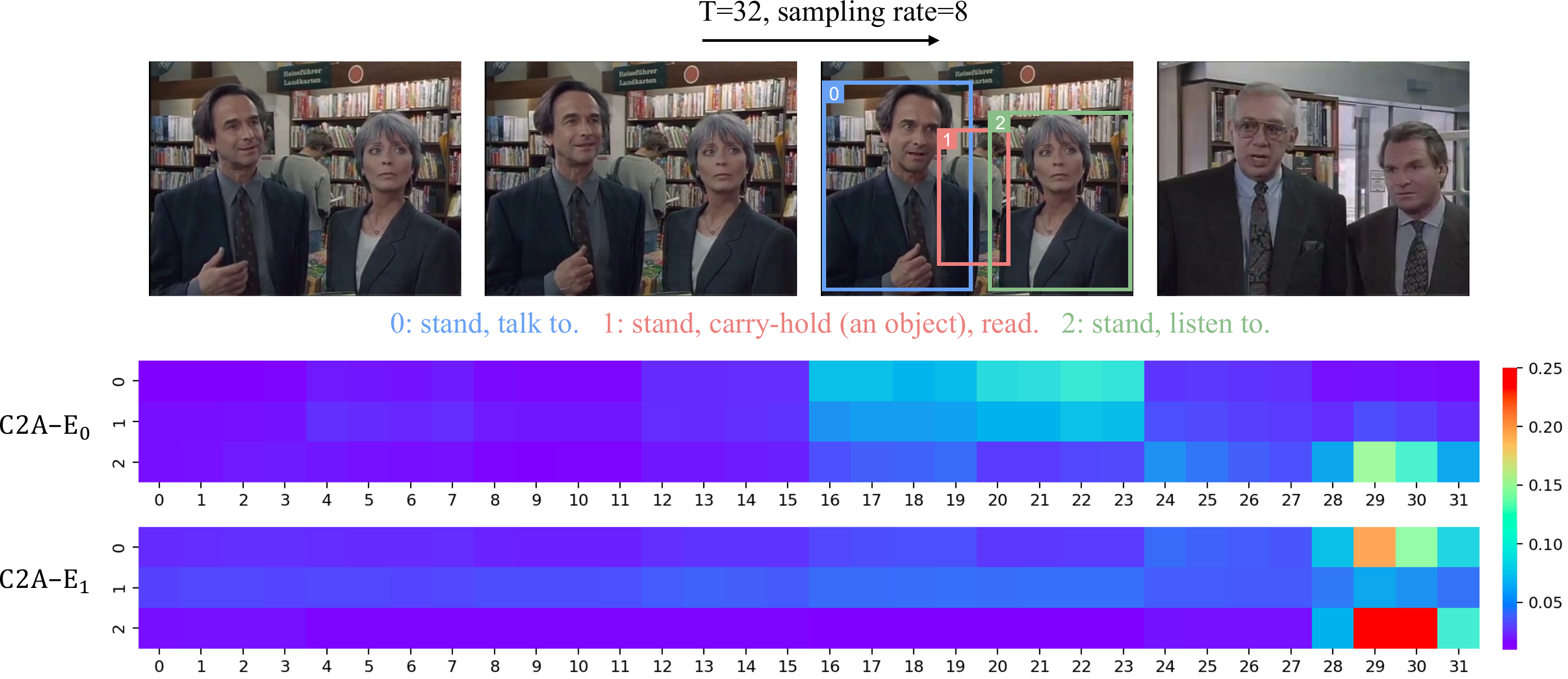}
			\end{minipage}
			\label{fig:temp_attn}
		}
		\end{center}
    \vspace{-5mm}
    \caption{\textbf{Visualization of A2C-R and C2A-E attention maps on AVA.} Best view in color.}
\vspace{-4mm}
\end{figure*}

\subsection{Qualitative results}
\label{sec:visualization}
To verify the effectiveness of actor-context modeling in CycleACR, we collect the visualizations of A2C-R and C2A-E.
For A2C-R, as shown in Fig.~\ref{fig:roi_attn}, we visualize the attention maps of the last A2C-R layer in the global branch for each actor to show how context aggregates actor features. 1st row: for each actor, the attention of the context is on the holding hand and the legs, respectively. 2nd row: the attention is on the watching eyes and the hand carrying the object. These visualizations indicate that the A2C-R module effectively captures actor region parts mostly related to actions and updates the actor-related context, as a better scene prior for follow-up modeling.

For C2A-E, as shown in Fig.~\ref{fig:temp_attn}, we visualize the attention maps of two C2A-E layers to show the diverse contribution of each frame to different actors of interest and further illustrate the effectiveness of preserving the temporal dimension. 
We observe that the attention for \textit{actor 0} and \textit{2} is mainly focused on the last quarter of the input clip, especially for \textit{actor 2}, as the location of the person she is listening to is of utmost importance. While \textit{actor 0} has an amount of attention around the keyframe in the first layer for capturing the motion of ``talk to''. For \textit{actor 1}, his attention evenly distributes around the input clip, as he only holds a book for reading and does not interact with the other people in the scene.
Maintaining temporal information of context for our cycle modeling allows each actor to consciously focus on different frames, removing the irrelevant cues to reduce difficulties and ambiguities caused by learning from a single global context. Also, stacking multiple CycleACR modules enables a rational division of labor and more efficient information exchange between actor and context. 
More visualizations are provided in Appendix~\S~\ref{sec:vis_supp}.

\section{Conclusion}
In this paper, we have presented a new perspective on actor-context modeling in videos for action detection. We introduce the core idea of cycle modeling to perform Actor-to-Context Reorganization and Context-to-Actor Enhancement. Based on this idea, we design the Cycle Actor-Context Relation network (CycleACR) and report the new start-of-the-art performance on two popular benchmarks of action detection. We hope the cycle modeling can serve as a general paradigm applicable to other relation modeling tasks for future research.

{\flushleft \bf Future work.} In this paper, we focus on effective relational modeling and propose the CycleACR module. We divide the action detection architecture into three parts: feature extraction, CycleACR relational modeling and instance interaction. The latter two parts belong to the detection head, and it is inefficient to perform them in serial. In future work, we will explore integrating these two parts into one unified module by extending the actor-context modeling to actor-`long-term context' modeling. Then, we would remove the instance interaction module and simplify the overall pipeline.


\appendix
\section*{Appendix}

\section{Pseudocode of CycleACR}
\label{sec:code}

\begin{algorithm}[ht!]
\vspace{-1mm}
\caption{PyTorch-like pseudocode of CycleACR.}
\label{algo:pseudo}
\algcomment{
\textbf{Notes}: \texttt{A2C-R} or \texttt{C2A-E} can be any attention instantiations, e.g., a modified non-local block, where \texttt{mm} is matrix multiplication and \texttt{k.t()} is \texttt{k}'s transpose.
}

\definecolor{codeblue}{rgb}{0.25,0.5,0.5}
\lstset{
  basicstyle=\fontsize{7pt}{7pt}\ttfamily\bfseries,
  commentstyle=\fontsize{7pt}{7pt}\color{codeblue},
  keywordstyle=\fontsize{7pt}{7pt},
}

\begin{lstlisting}[language=python]
# a_local: actor local feature cropped from video feature
# map via RoIAlign
# c_local: temporal context generated by Pre-Processor
# on video feature map

a = maxpool(a_local)  # spatial max pooling
a_local = concat([a, a_local])
c_global = avgpool(c_local)  # temporal average pooling

# local branch of CycleACR
for _ in range(N):  # number of A2C-R blocks
    c_local = A2C-R(c_local, a_local)

out_local = a
c_local = c_local + pos  # add temporal position encoding
for _ in range (N):  # number of C2A-E blocks
    out_local = C2A-E(out_local, c_local)

# global branch of CycleACR
for _ in range (N):
    c_global = A2C-R(c_global, a_local)
out_global = c_global 

# combine outputs of two branches
out_final = concat([out_local, out_global])

# actor-to-context reorganization
def A2C-R (q, m):
    q_, k, v = w_q(q), w_k(m), w_v(m)
    attn = softmax(mm(q_, k.t()) * scale)
    out = mm(attn, v)
    out = q + dropout(w_out(relu(norm(out))))
    return out 

# context-to-actor enhancement
def C2A-E(q, m):
    return A2C-R(q, m)
\end{lstlisting}
\vspace{-1mm}
\end{algorithm}

In Alg.~\ref{algo:pseudo}, we include pseudocode for the core implementation of the CycleACR module.
We perform cycle modeling using two parallel branches: the local branch and the global branch, which consists of two components, A2C-R and C2A-E.
We can use different attention mechanisms as A2C-R/C2A-E instantiation, e.g., modified non-local~\cite{wang2018non}, transformer~\cite{DBLP:conf/nips/VaswaniSPUJGKP17} and dynamic convolution~\cite{DBLP:conf/cvpr/SunZJKXZTLYW021}. Here we take the non-local one as an illustration.

\section{Result Analysis and Visualization}
\label{sec:vis_supp}

\begin{figure*}[t]
\centering
\includegraphics[width=1.0\linewidth]{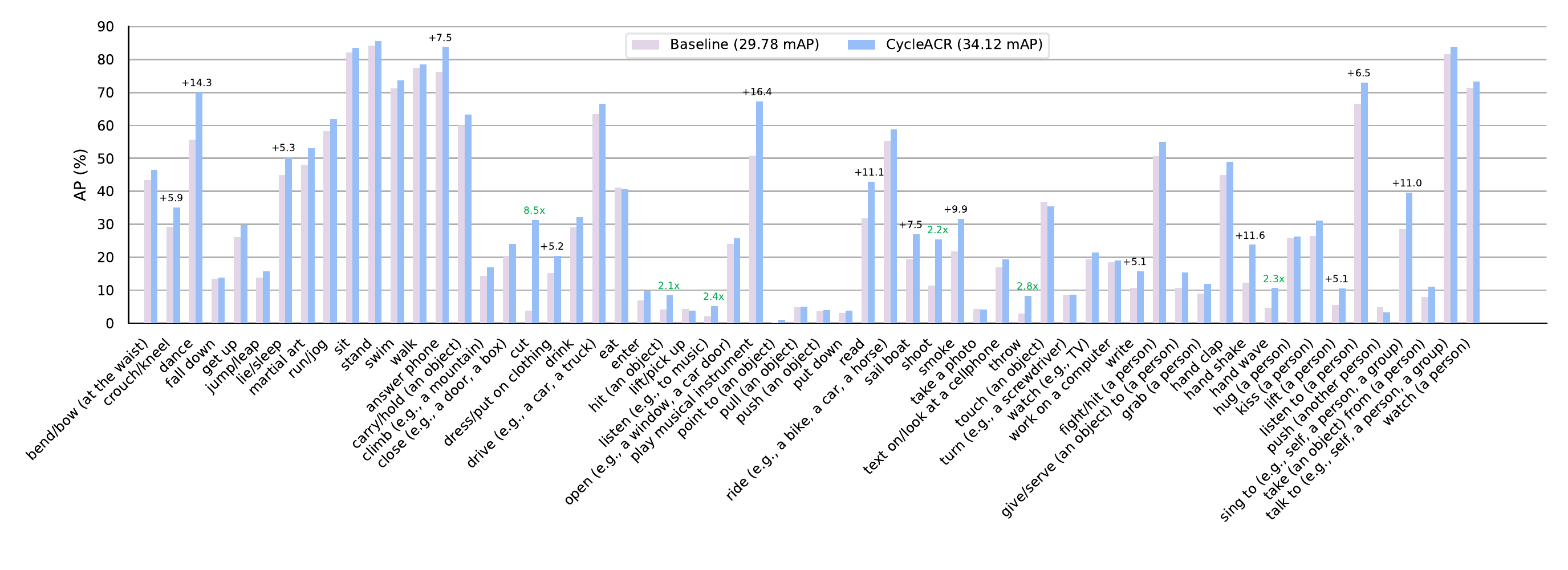}
\vspace{-6mm}
\caption{\textbf{Per-category AP for a SlowFast baseline(29.78 mAP) and the proposed CycleACR (34.12 mAP) on AVA.} Categories that increase in absolute value by more than 5\% are marked in \textbf{\textcolor[RGB]{0,0,0}{black}} and those more than twice the AP of the baseline are marked in \textbf{\textcolor[RGB]{0,176,80}{green}}.}
\label{fig:per_ap}
\end{figure*}

Firstly, we compare the per-category performance of a SlowFast~\cite{DBLP:conf/iccv/Feichtenhofer0M19} baseline and our proposed CycleACR on AVA~\cite{DBLP:conf/cvpr/GuSRVPLVTRSSM18}, as shown in Fig.~\ref{fig:per_ap}. Our method improves in 55 out of 60 categories, especially for those who require reasonable context modeling, e.g., {\it ``cut''} ($8.5\times$), {\it ``play musical instrument''} ($+16.4\%$), and {\it ``read''} ($+11.1\%$).

\begin{figure*}[t]
\centering
\includegraphics[width=0.95\linewidth]{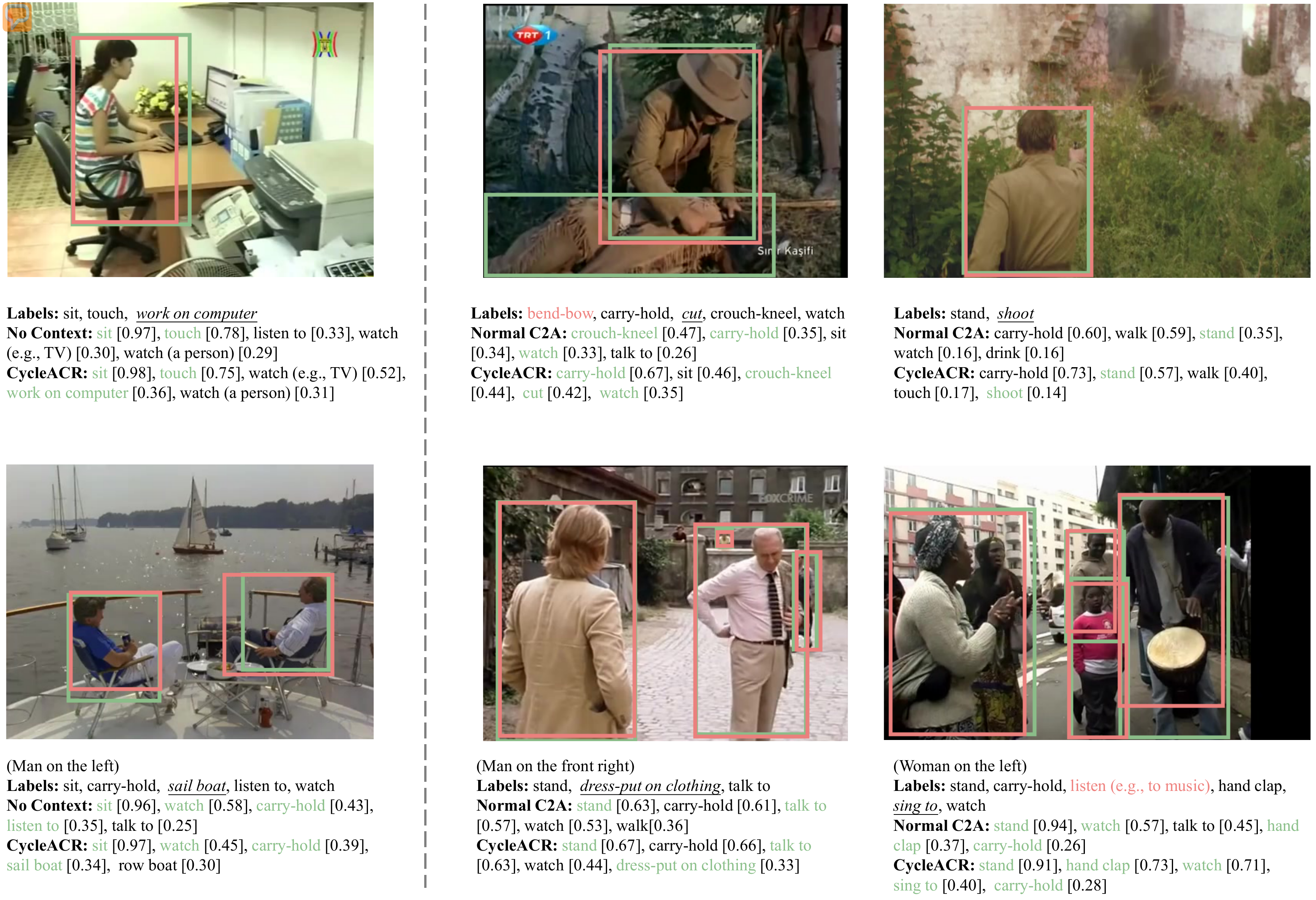}
\vspace{-2mm}
\caption{\textbf{Example results on AVA.} In the images, ground-truth boxes are shown in \textbf{\textcolor[RGB]{140,192,139}{green}} and detected boxes are shown in \textbf{\textcolor[RGB]{239,130,127}{red}}. The text below images shows the ground-truth labels and top-5 predictions from models on the given actor. The first column shows a comparison of the results between no context modeling and CycleACR, while the last two columns are between normal context-to-actor (C2A) modeling and CycleACR. The actions not predicted by both models are marked in \textbf{\textcolor[RGB]{239,130,127}{red}} and those successfully predicted are marked in \textbf{\textcolor[RGB]{140,192,139}{green}}. Those actions successfully predicted by CycleACR but failed by the other model are marked in {\it italics} with \underline{underline}.
}
\label{fig:preds}
\vspace{-3mm}
\end{figure*}

\begin{figure*}[t]
\centering
\vspace{-5mm}
\includegraphics[width=0.9\linewidth]{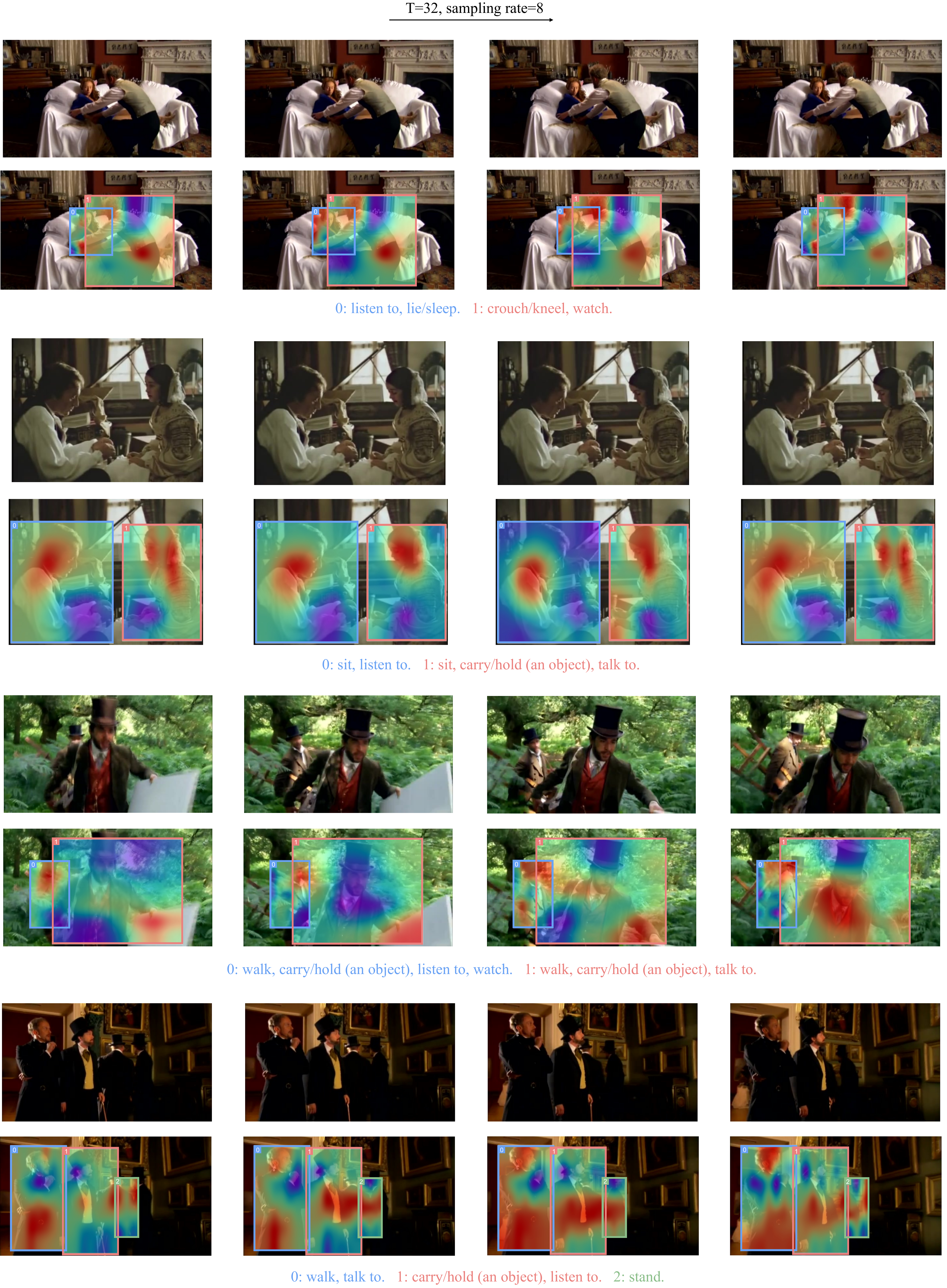}
\caption{\textbf{More visualizations of A2C-R attention maps on AVA.} For each example, 1st row: RGB frames of the input clip with a stride of 8 for viewing better, where the third image represents the keyframe. 2nd row: attention weights of the last A2C-R layer between each temporal context and different actors of interest.}
\label{fig:roi_attn_supp}
\vspace{-3mm}
\end{figure*}

\begin{figure*}[t]
\centering
\vspace{-5mm}
\includegraphics[width=0.75\linewidth]{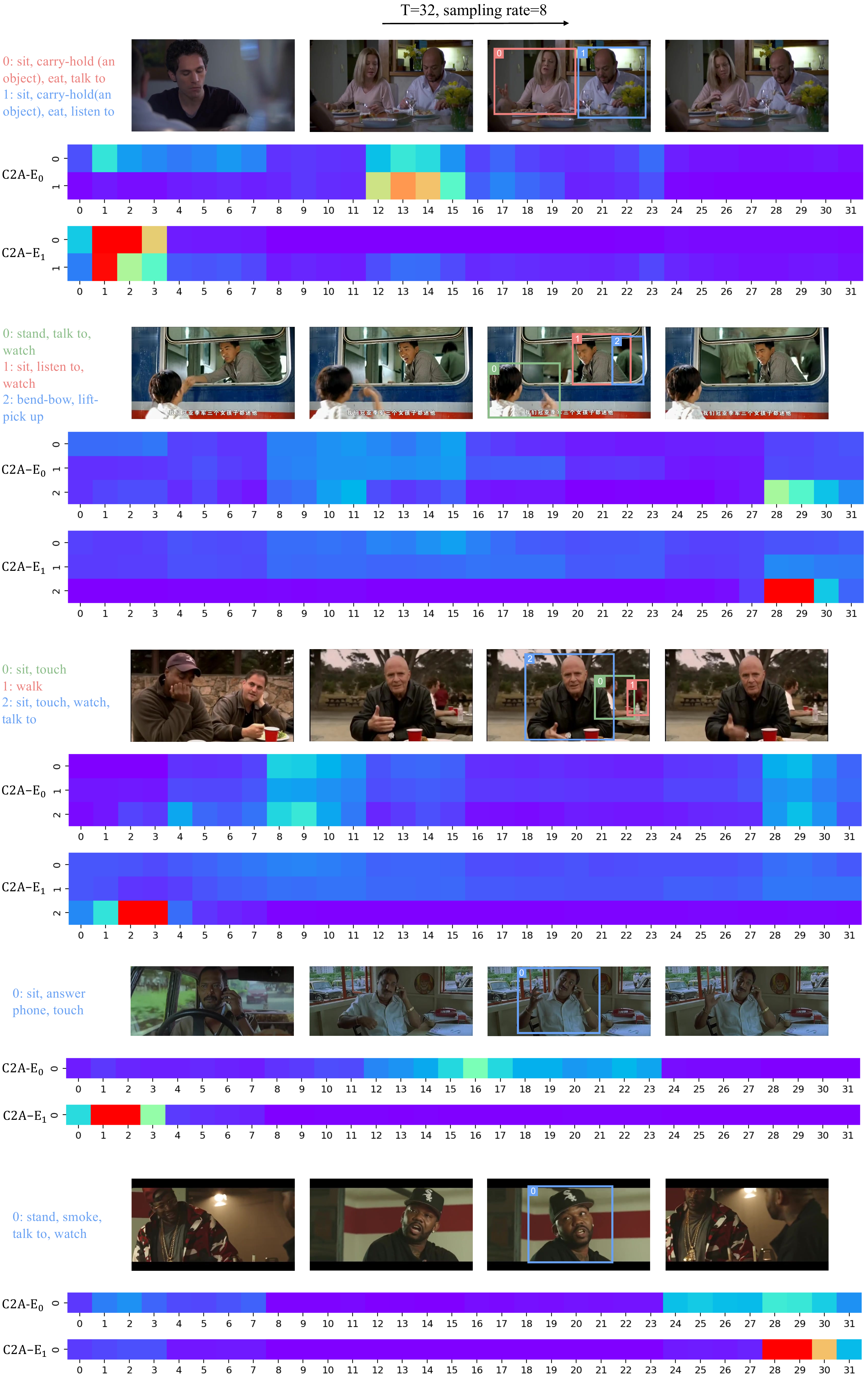}
\caption{\textbf{More visualizations of C2A-E attention maps on AVA.} For each example, 1st row: RGB frames of the input clip with a stride of 8 for viewing better, where the third image represents the keyframe. 2nd and 3rd row: attention weights of the $i^{th}$ C2A-E layer between actors of interest on the keyframe and 32-frame context.}
\label{fig:temp_attn_supp}
\vspace{-3mm}
\end{figure*}

Then, we collect more visualizations of A2C-R and C2A-E attention maps as a supplementary of Figure 6 in our paper, as shown in Fig.~\ref{fig:roi_attn_supp} and Fig.~\ref{fig:temp_attn_supp}. 

For A2C-R, we have visualized the attention maps of A2C-R in the global branch, and we supplement the A2C-R visualization results in the local branch to show the aggregation effect of each local context, as shown in Fig.~\ref{fig:roi_attn_supp}.
We observe that with slow actor movement, the attention of each frame is similar, as in {\it examples 1}. 
The benefits of refining actor-related context separately on each frame are twofold: (1) integrating the reorganized context of multiple frames to enrich action semantics, e.g., in {\it example 2}, only the frame on the third column captures the hand information associated with the action ``carry/hold (an object)''; (2) reducing the interference between frames in cases with large movement or camera translation, e.g., in {\it examples 3} and {\it 4}, partial actor features deviate from the bounding boxes in other frames.

For C2A-E, we visualize more attention maps of two C2A-E layers in the local branch, as shown in Fig.~\ref{fig:temp_attn_supp}.
We observe that the first layer is more focused on effective information of temporal context around the keyframe, i.e., capturing action cues more relevant to the target actor. While the second layer of attention weights is differentiated between actors, capturing distinct scene information for their action recognition, e.g., the people they talk to in {\it examples 1} and {\it 3}, the object being lifted in {\it example 2}, the person on the phone in {\it example 4}, and the cigarette in {\it example 5}. It demonstrates the importance of maintaining temporal information of context and the effectiveness of stacking multiple CycleACR modules.

Finally, by comparing the prediction results of CycleACR with no context modeling and normal context-to-actor modeling, we demonstrate that cycle modeling can better understand actor-context relations, as shown in Fig.~\ref{fig:preds}. 
The first column shows the predictions of no context modeling and CycleACR, where interactive objects (e.g., computer, boat) are outside of human bounding boxes and require actor-context modeling to capture such scene information. 
The last two columns show the predictions of normal context-to-actor modeling and CycleACR. In the first row, the context-to-actor fails to recognize the details of hand movements (e.g., {\it ``cut''}, {\it``shoot''}), which are overwhelmed by the amount of irrelevant scene information. Instead, the cycle modeling consciously removes action-irrelevant information to capture these essential details easier.
In the second row, the unidirectional context-to-actor misunderstands {\it ``dress-put on clothing''} as {\it ``carry-hold clothing''} and {\it ``sing to''} as {\it ``talk to''}, both of which are the consequence of an inadequate understanding of context. In contrast, CycleACR can deduce the correct action by modeling the high-order relation between the posture of the person and the clothing held, the singer and the percussionist.

Note that actions marked in \textbf{\textcolor[RGB]{239,130,127}{red}} in Fig.~\ref{fig:preds} indicate actions missed by both models, which suggests that there is still much room for future improvement in relation modeling.

{\small
\bibliographystyle{ieee_fullname}
\bibliography{ref}
}

\end{document}